\documentclass[conference]{IEEEtran}
\IEEEoverridecommandlockouts
\usepackage{cite}
\usepackage{amsmath,amssymb,amsfonts}
\usepackage{algorithmic}
\usepackage{graphicx}
\usepackage{textcomp}
\usepackage{xcolor}
\def\BibTeX{{\rm B\kern-.05em{\sc i\kern-.025em b}\kern-.08em
    T\kern-.1667em\lower.7ex\hbox{E}\kern-.125emX}}
    
\usepackage[hyphens]{url}
\usepackage[breaklinks,hidelinks]{hyperref}
\usepackage[utf8]{inputenc}
\usepackage{graphicx}

\usepackage{subcaption}
\usepackage{natbib}
\usepackage{cleveref}
\usepackage[acronym]{glossaries}
\usepackage{svg}

\usepackage{booktabs}
\usepackage{multirow}

\newcommand{\n}[0]{58,672}

\makeglossaries
\newacronym{dpm}{DPM}{Deformable Parts Model}
\newacronym{mil}{MIL}{Multiple Instance Learning}
\newacronym{wsod}{WSOD}{Weakly Supervised Object Detection}

\newacronym{ai}{AI}{artificial intelligence}

\newglossaryentry{pa}{
    name={People-Art},
    description={DEF}
}
\newacronym{sin}{SIN}{Stylized-ImageNet}
\newglossaryentry{sc}{
    name={StyleCOCO},
    description={DEF}
}

\newacronym{map}{mAP}{mean average precision}
\newacronym{iou}{IoU}{intersection over union}
\newglossaryentry{ap}{
    name={AP},
    description={COCO mAP metric}
}
\newglossaryentry{ap5}{
    name={\ensuremath{\textrm{AP}^{50}}},
    description={COCO mAP metric at 50\% overlap}
}
\newglossaryentry{ap75}{
    name={\ensuremath{\textrm{AP}^{75}}},
    description={COCO mAP metric at 75\% overlap}
}

\newacronym{nn}{NN}{Neural Network}
\newacronym{cnn}{CNN}{Convolutional Neural Network}
\newacronym{rcnn}{R-CNN}{Region Based Convolutional Neural Network}
\newacronym{frcnn}{Fast R-CNN}{Fast Region Based Convolutional Neural Network}
\newacronym{frrcnn}{Faster R-CNN}{Faster Region Based Convolutional Neural Network}

\newglossaryentry{rn50}{
    name={ResNet-50},
    description={}
}
\newglossaryentry{rn152}{
    name={ResNet-152},
    description={}
}

\newacronym{sgd}{SGD}{stochastic gradient descent}

\newglossaryentry{exp_ntrain}{
    name={Experiment 1},
    description={DEF}
}
\newglossaryentry{exp_ap}{
    name={Experiment 2},
    description={DEF}
}

\begin{document}

\title{Improving Object Detection in Art Images Using Only Style Transfer
\thanks{The research presented here has received funding from the European Union’s Horizon 2020 research and innovation programme under grant agreement No 727040 (the GIFT project).}
}

\author{\IEEEauthorblockN{David Kadish}
\IEEEauthorblockA{\textit{Digital Design Department} \\
\textit{IT University of Copenhagen}\\
Copenhagen, Denmark \\
davk@itu.dk}
\and
\IEEEauthorblockN{Sebastian Risi}
\IEEEauthorblockA{\textit{Digital Design Department} \\
\textit{IT University of Copenhagen}\\
Copenhagen, Denmark \\
sebr@itu.dk}
\and
\IEEEauthorblockN{Anders Sundnes Løvlie}
\IEEEauthorblockA{\textit{Digital Design Department} \\
\textit{IT University of Copenhagen}\\
Copenhagen, Denmark \\
asun@itu.dk}
}

\maketitle

.
\begin{abstract}
Despite recent advances in object detection using deep learning neural networks, these neural networks still struggle to identify objects in art images such as paintings and drawings.
This challenge is known as the cross depiction problem and it stems in part from the tendency of neural networks to prioritize identification of an object's texture over its shape.
In this paper we propose and evaluate a process for training neural networks to localize objects --- specifically people --- in art images.
We generate a large dataset for training and validation by modifying the images in the COCO dataset using AdaIn style transfer.
This dataset is used to fine-tune a Faster R-CNN object detection network, which is then tested on the existing \gls{pa} testing dataset.
The result is a significant improvement on the state of the art and a new way forward for creating datasets to train neural networks to process art images.
\end{abstract}
\section{Introduction}

Image recognition systems --- especially ones based on deep neural networks --- have improved significantly in recent years~\citep{lecun_deep_2015}, yet they are often brittle and struggle if images are grainy or noisy.
More worryingly, they can be manipulated into making incorrect classifications by malicious pixel changes, sometimes on as little as a single pixel~\citep{su_one_2019}.
Neural networks struggle even more to recognise objects depicted in different styles such as drawings or paintings, something the human mind can do readily even as a small child.
This is known as the cross-depiction problem~\citep{Cai2015,boulton_artistic_2019}.

Meanwhile, in the museum sector there is a growing interest in the use of object detection in artworks~\citep{ciecko2020}, partly for the purpose of improving metadata in order to better support search, e.g. by making it possible to search for visual motifs; and partly in order to set up new creative possibilities for artists and designers, to create innovative art experiences e.g. through apps such as \textit{Smartify} or \textit{Send Me SFMoma}~\citep{mollica_send_2017}.
As is the case with many AI problems, a major challenge for improving object detection in artworks is the lack of large, high quality datasets for training. 

In this paper, we introduce a technique inspired in part by \citet{Geirhos2018}
to generate a dataset and train an object detector to locate people in artworks.
\citet{Geirhos2018} used style transfer to demonstrate that neural networks prioritize texture over shape in classifying images and to train a neural network to be more robust to textural distortions such as noise.
Here, style transfer is used to address the cross-depiction problem.
We create a large training dataset (\n{} images) of artistically-styled images from already-labelled photographs, train an object detector, and test it on the artwork dataset People-Art~\citep{Cai2015}, which results in a significant improvement on past detection efforts on this dataset.


This research contributes to the field of computer vision by expanding the visual domains available for object detection, for instance making it possible to search for motifs in non-photographic databases. 
Furthermore, it is a step towards developing vision algorithms that have more general representations of objects that resemble the ways humans see objects, allowing for more robust object detection in photos and video (e.g. algorithms that can better handle noise and distortions).

Improving object detection will enable new ways to search and study particular objects in art collections.
This may enable new strategies for doing quantitative research on art history and cultural heritage in the field of digital humanities \citep{berry_digital_2017,manovich_ai_2018,manovich_cultural_2020}.
In combination with existing metadata this would make it possible to search for specific depictions of people, for instance based on the colour of their skin, in order to quantitatively explore depictions of race in art (searching for images based on colour profile is already a well-established technique).
Another use case might be to search the collections of a national museum for depictions of objects from other countries to explore international influences on culture in different historical eras.
Such metadata could enable the design of more engaging experiences with museum collections, e.g. by giving users the means to search for objects relating to their personal interests \citep{ciecko2020,Merritt2017,Bailey2019,Smith2019}.
\section{Background}

A perpetual problem in \gls{ai} is that the creation of large-scale, high-quality datasets is time-consuming and expensive~\citep{Chang2017}.
For some tasks an effort has been made to compile a canonical dataset that can be widely used by the \gls{ai} community~\citep{Deng2010,Lin2014}; for less common tasks the challenge persists.


\begin{table}[t]
\centering
\caption{Labelled art image datasets}
\begin{tabular}{lrrrl} \toprule
\textbf{Name}      & \textbf{Images} & \textbf{Labels} & \textbf{Classes} & \textbf{Task}    \\ \midrule
Paintings Database & 10000           & 8629            & 10                  & Classification   \\
BAM                & 2500000         & 393000          & n/a                 & Classification   \\
Photo-Art-50       & $\sim$5000      & $\sim$5000      & 50                  & Classification   \\
PACS               & 9991            & n/a             & 7                   & Classification   \\
People-Art         & 4631            & 3487            & 1                   & Obj. Detection   \\ \bottomrule    
\end{tabular}
\label{tab:art_dbs}
\end{table}

The major datasets for object detection training are comprised mainly of photographs.
This leads to the cross-depiction problem~\citep{Cai2015}, where \glspl{nn} struggle to correctly identify objects depicted in different styles --- they may be able to pick out a car in a photograph, but struggle with a pencil sketch of the same vehicle.
\citeauthor{Cai2015} attempt to tackle this problem using neural networks pre-trained on a large database of photographs and fine-tuned on a smaller dataset of labelled artwork.
Achieving a maximum \gls{ap5} score of 0.4, they conclude that deep learning has yet to demonstrate that it is able to solve the cross-depiction problem.

One of the potential reasons for this is that neural networks are not good at recognizing shapes.
They base their decisions mainly on the texture of the objects that they are examining~\citep{Geirhos2018}.
While neural networks tend to be biased towards textures, humans tend to exhibit a strong bias towards shape~\citep{Geirhos2018}.
Artistic representations tend to also be shape-biased, making artwork especially difficult for most neural networks to process.

\subsection{Datasets}
The problem of detecting objects in artwork has proven to be particularly challenging.
For image classification --- categorizing the overall content of an image --- a large training set can be created using metadata like image titles, descriptions, and artist- and crowd-generated tags, as in the Behance Artistic Media (BAM) dataset ~\citep{Wilber2017}.
Object detection, however, requires training data with labelled bounding boxes which indicate where in an image an object is localized, and few sources of this type of data exist (see \cref{tab:art_dbs}).

The main data source used for object detection in artwork is the \gls{pa} dataset, originally described by \citet{Cai2015}.
\Gls{pa} consists of a set of images of artwork in which people have been labelled with bounding boxes. The dataset includes a training set, a validation set and a test set.
\Cref{tab:pa_stats} shows the breakdown of total images, positive images (containing people) and labelled people in the dataset.
The dataset encompasses a wide range of artistic styles, but is notably small for training a deep neural network.

\begin{table}[t]
\caption{Art image object detection datasets}
\begin{center}
\begin{tabular}{llrrr} \toprule
\textbf{Dataset}  & \textbf{Subset} & \textbf{Images} & \textbf{Positive} & \textbf{People} \\ \midrule
\multirow{3}{*}{People-Art} & training & 1,627 & 521 & 1,324 \\
& validation & 1,387 & 442 & 1,080 \\
& testing & 1,617 & 520 & 1,083 \\
\\
\multirow{2}{*}{StyleCOCO} & training & 58,672 & 58,672 & 239,845 \\
& validation & 2,688 & 2,688 & 10,997 \\
\bottomrule
\end{tabular}
\label{tab:pa_stats}
\end{center}
\end{table}

\subsection{Object detection}
Despite the limited availability of training data, a number of efforts have been made to detect people using deep neural networks --- and a variety of other strategies --- in the \gls{pa} test set.
\citet{Cai2015} used \glspl{rcnn} pre-trained on ImageNet and attempted fine-tuning them on the VOC2007 dataset and \gls{pa}.
\citet{Westlake2016} produced a similar attempt, using \gls{frcnn} with a variety of architectures also pre-trained on ImageNet and fine-tuned on VOC 2007 or \gls{pa}.
\citet{Redmon2016} also tested the YOLO network on \gls{pa}, in order to see how well it performed in a cross-depiction test.

Evaluating these efforts is non-trivial, but the standard metrics are a set of average precision and recall measurements that assess the performance of the detector at different levels of bounding box overlap with the labelled box.
The classic metric was devised as part of the PASCAL VOC challenge (Everingham et al. 2010) and is the \gls{map} with a 50\% overlap between bounding boxes.
The results of these trials using this metric are shown in \cref{tab:aps_from_others} and \cref{fig:ap_compare}.
The \gls{frcnn} network using a VGG16 backbone, pre-trained on ImageNet and fine-tuned on the \gls{pa} training set performed the best, scoring 0.58 in the \gls{ap5} test.

Is this a good result?
It can be difficult to know what to compare this to.
Obviously, a score of 1.0 would be desirable, but likely unrealistic.
The best score for person detection in photographs in the latest COCO detection challenge was 0.85\footnote{From \url{https://cocodataset.org/\#detection-leaderboard}, this is the Chal17 score for \texttt{person-person} category by \texttt{Megvii (Face++)}.}, but this detector was trained to look for objects in 80 different categories --- a much more difficult task than just searching for people.
Clearly there is room for improvement.
\section{Dataset Generation using Style Transfer}
%

Given the assertion of \citet{Geirhos2018} that \glspl{cnn} trained on ImageNet were generally texture-biased, we first tried using their \acrlong{sin}-trained \gls{rn50} network as a backbone for a \gls{frrcnn} object detector to see if it would improve over a backbone trained on ImageNet.
This failed to yield a significant improvement, but the process inspired an alternative use of style transfer.
\citet{Geirhos2018} had applied a style transfer to the images from ImageNet to generate a set of images they call \gls{sin}.
If we could use a similar process on a database of images that was already labelled for object detection, we could create a massive, pre-labelled dataset of images that could be used for object detection in art images.

\subsection{Style transfer}
AdaIn style transfer~\citep{Huang2017} is a neural network that takes two input images and applies the style of one to the content of the other.
It is the mechanism used in \citet{Geirhos2018} to create the stylized dataset that they call \acrlong{sin}.

We follow the basic process described in \citet{Geirhos2018}, using the Painter by Numbers dataset from Kaggle\footnote{\url{https://www.kaggle.com/c/painter-by-numbers/} (accessed 14 Oct 2020).} as the source of stylized images, modifying each input image with a single style with a stylization weight of 1 and not changing the size or crop of the input files.
The stylization script used is from \url{https://github.com/bethgelab/stylize-datasets}.

For the source images, we chose the COCO 2017 dataset --- a library of 123,287 photographs with bounding box object detection labels~\citep{Lin2014}.
The resulting dataset is referred to as StyleCOCO.

\subsection{StyleCOCO}

\newcommand\imggridwidth{0.19\textwidth}
\begin{figure*}[h]
     \centering
     \begin{subfigure}[b]{\imggridwidth}
         \centering
         \includegraphics[width=\textwidth]{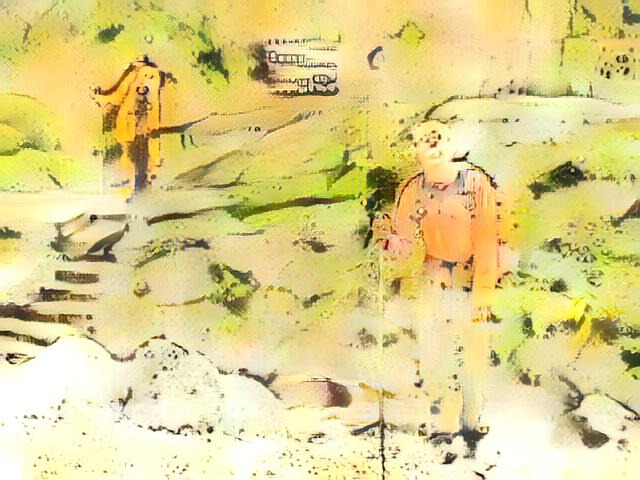}
         \\
         \includegraphics[width=\textwidth]{images/examples/000000032887_original.jpg}
         \centering
     \end{subfigure}
     \hfill
     \begin{subfigure}[b]{\imggridwidth}
         \centering
         \includegraphics[width=\textwidth]{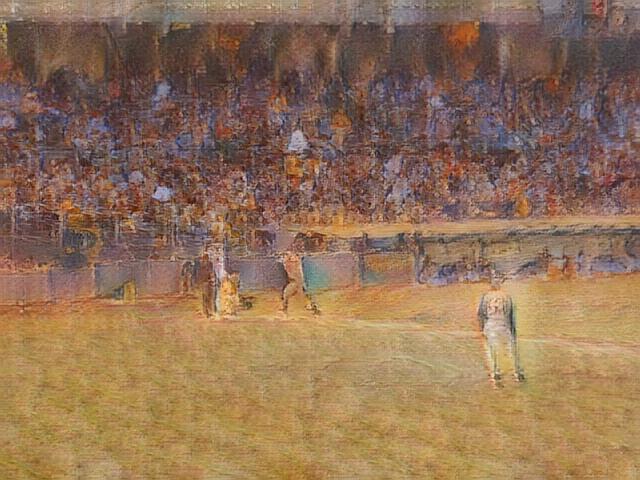}
         \\
         \includegraphics[width=\textwidth]{images/examples/000000065798_original.jpg}
         \centering
     \end{subfigure}
     \hfill
     \begin{subfigure}[b]{\imggridwidth}
         \centering
         \includegraphics[width=\textwidth]{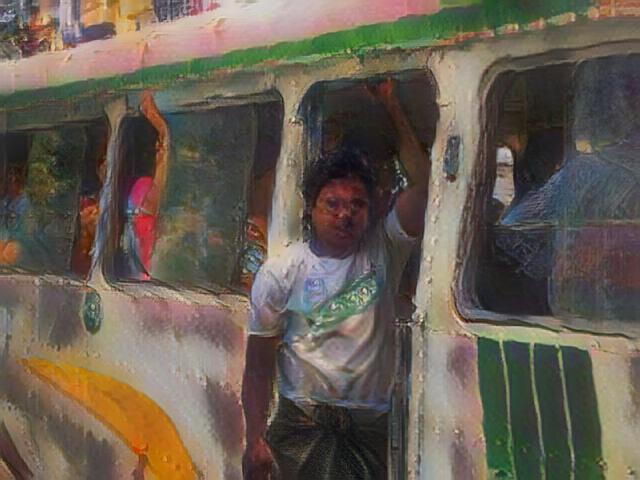}
         \\
         \includegraphics[width=\textwidth]{images/examples/000000319534_original.jpg}
         \centering
     \end{subfigure}
     \hfill
     \begin{subfigure}[b]{\imggridwidth}
         \centering
         \includegraphics[width=\textwidth]{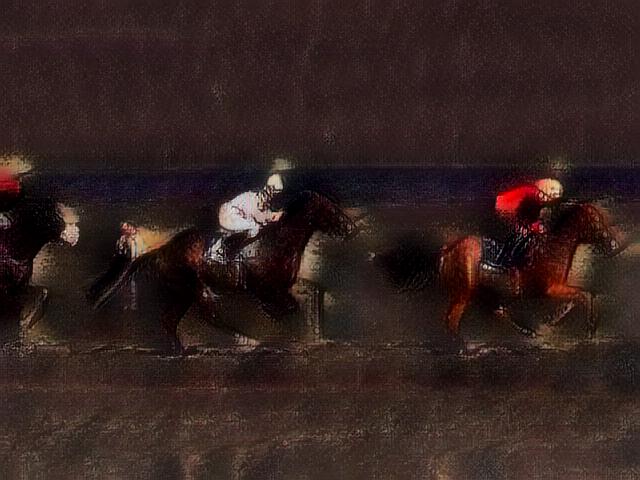}
         \\
         \includegraphics[width=\textwidth]{images/examples/000000360661_original.jpg}
         \centering
     \end{subfigure}
     \hfill
     \begin{subfigure}[b]{\imggridwidth}
         \centering
         \includegraphics[width=\textwidth]{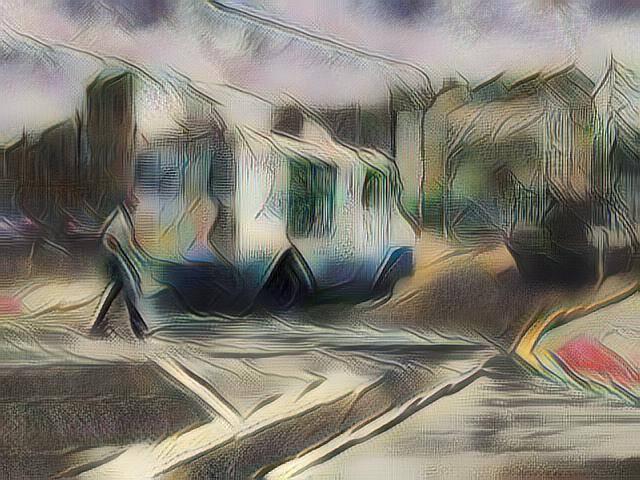}
         \\
         \includegraphics[width=\textwidth]{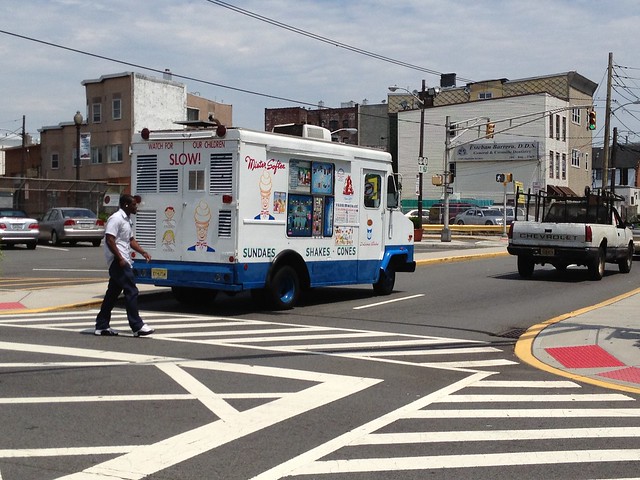}
         \centering
     \end{subfigure}
     \newline
     \newline
     \begin{subfigure}[b]{\imggridwidth}
         \centering
         \includegraphics[width=\textwidth]{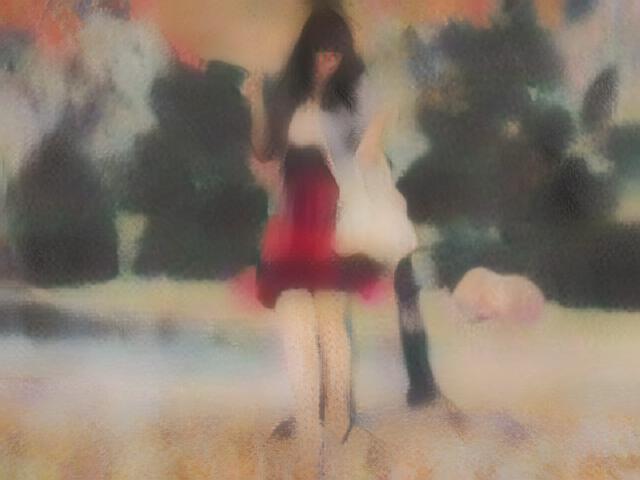}
         \\
         \includegraphics[width=\textwidth]{images/examples/000000401446_original.jpg}
         \centering
     \end{subfigure}
     \hfill
     \begin{subfigure}[b]{\imggridwidth}
         \centering
         \includegraphics[width=\textwidth]{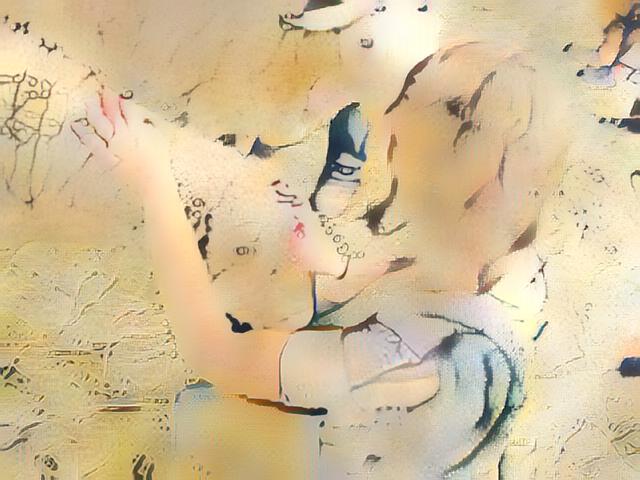}
         \\
         \includegraphics[width=\textwidth]{images/examples/000000458755_original.jpg}
         \centering
     \end{subfigure}
     \hfill
     \begin{subfigure}[b]{\imggridwidth}
         \centering
         \includegraphics[width=\textwidth]{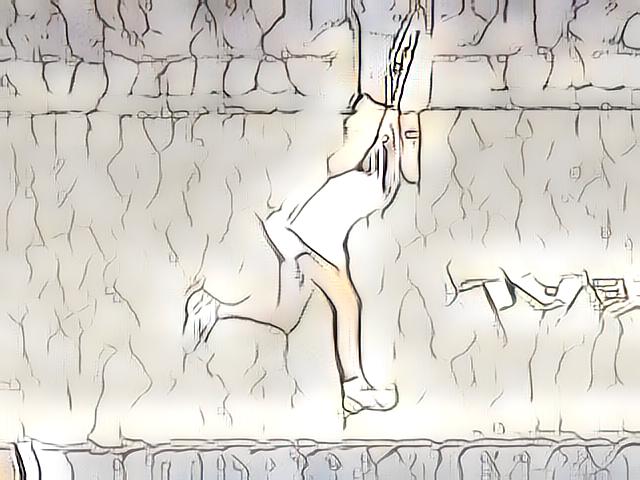}
         \\
         \includegraphics[width=\textwidth]{images/examples/000000475191_original.jpg}
         \centering
     \end{subfigure}
     \hfill
     \begin{subfigure}[b]{\imggridwidth}
         \centering
         \includegraphics[width=\textwidth]{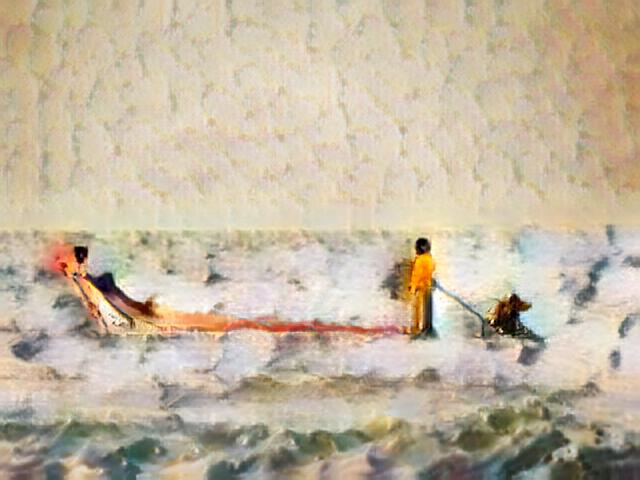}
         \\
         \includegraphics[width=\textwidth]{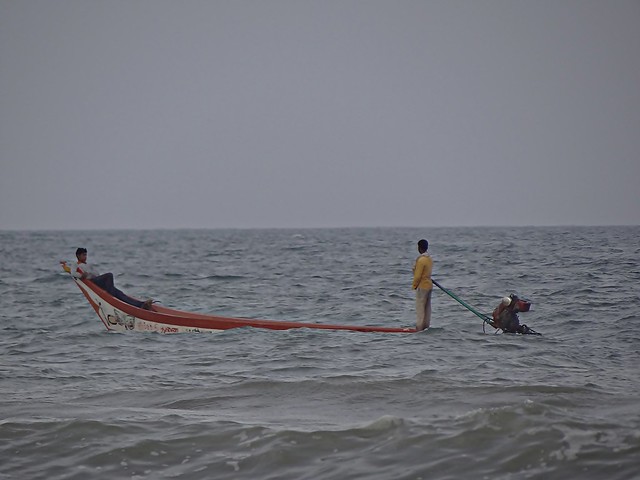}
         \centering
     \end{subfigure}
     \hfill
     \begin{subfigure}[b]{\imggridwidth}
         \centering
         \includegraphics[width=\textwidth]{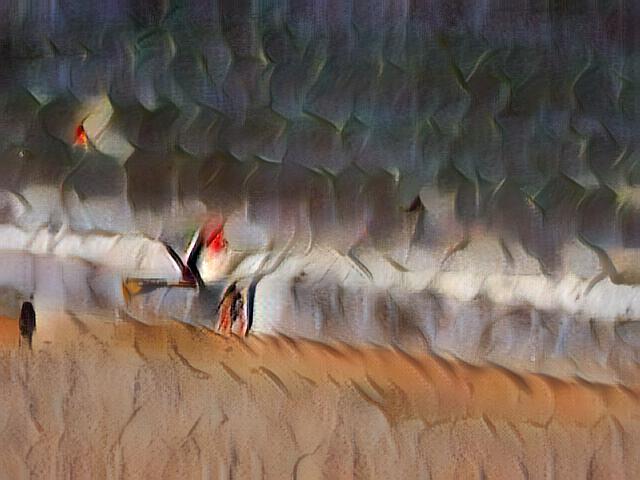}
         \\
         \includegraphics[width=\textwidth]{images/examples/000000540932_original.jpg}
         \centering
     \end{subfigure}
     \hfill
        \caption{Sample images from the StyleCOCO dataset (above) paried with the original image from the COCO dataset (below)}
        \label{fig:samples}
\end{figure*}

StyleCOCO is a subset of the original COCO dataset~\citep{Lin2014} with images stylized to look like artwork using AdaIn style transfer.
It consists of a training set of \n{} images and a validation set of 2,688 images\footnote{The set of images is smaller than the full COCO dataset because we excluded images that did not contain a labelled person.}.
The images are annotated for object detection for the person class and bounding box and segmentation information is retained.
A total of 239,845 annotated people are found in the training set and a further 10,997 are found in the validation set.

Images from the original COCO dataset containing no people were removed from the dataset as the pixels not containing people provide sufficient negative examples for neural network training.
No testing set was stylized as the dataset was only intended for training a neural network to be tested on data from the \gls{pa} dataset.
A breakdown of the number of images and labelled people is shown in \cref{tab:pa_stats} and sample images from the StyleCOCO dataset are shown in \cref{fig:samples}.
\section{Model Training}
The models employed in this study were trained using the \texttt{PyTorch} library and its built-in \texttt{FasterRCNN} model with a \gls{rn152} backbone pre-trained on ImageNet.
Models were fine-tuned on the StyleCOCO training dataset and evaluated during training using a random subset of 2,000 images from the StyleCOCO validation dataset.
Early stopping was engaged to prevent overfitting of models to the training data.
After fine-tuning was complete, models were tested on the full \gls{pa} test dataset.
The reported AP scores are from this testing phase.

The code for creating the dataset and training and testing the model is available at \url{https://github.com/dkadish/Style-Transfer-for-Object-Detection-in-Art}.

\subsection{Experiments}
Two experiments were conducted to test the validity of using stylized photographs to fine-tune a neural network for object detection in artwork.
In the first (\gls{exp_ntrain}), we tested the impact of the number of training images on the performance of the object detector.
We trained 7 models with exponentially\footnote{Following the exponential pattern, the final experiment should have had 64000, but due to the limitations of the COCO dataset there were only \n{} images available, so the entire dataset was used.} increasing amounts of input data, starting at 1000 images.
Images were randomly selected from the full dataset for each model and only those images were used in training.
The models were then tested on the \gls{pa} testing dataset.
These results were compared against each other in order to test the impact of additional training data on model performance.

The second experiment (\gls{exp_ap}) compares performance of a model fine-tuned on StyleCOCO on the \gls{pa} testing dataset with other results reported in the literature.
This experiment establishes the performance of the method detailed here against the state-of-the-art in object detection in artwork.

It is important to note that in both of these experiments, the testing results are achieved without having to use any labelled data from target distribution in training.
The \gls{pa} training datasets are not used in any way to train the \gls{nn} that eventually detects objects in the \gls{pa} testing set.

\subsection{Model and training parameters}
The model used in both experiments is a \gls{frrcnn} network with a \gls{rn152} backbone pre-trained on ImageNet.
Models were fine-tuned using the StyleCOCO training and validation datasets --- or a subset thereof in the case of \gls{exp_ntrain}.
Default training parameters from \texttt{PyTorch} were used except where modifications were shown to improve performance and those non-default parameters are listed below.
2 layers of the backbone were made trainable and the remaining 3 were frozen to retain the pre-trained ImageNet weights. 
Models were trained for 15 epochs, though early stopping (\texttt{patience}=3.0) was employed in order to avoid overfitting.

\begin{figure}
    \centering
    \includegraphics[width=\linewidth]{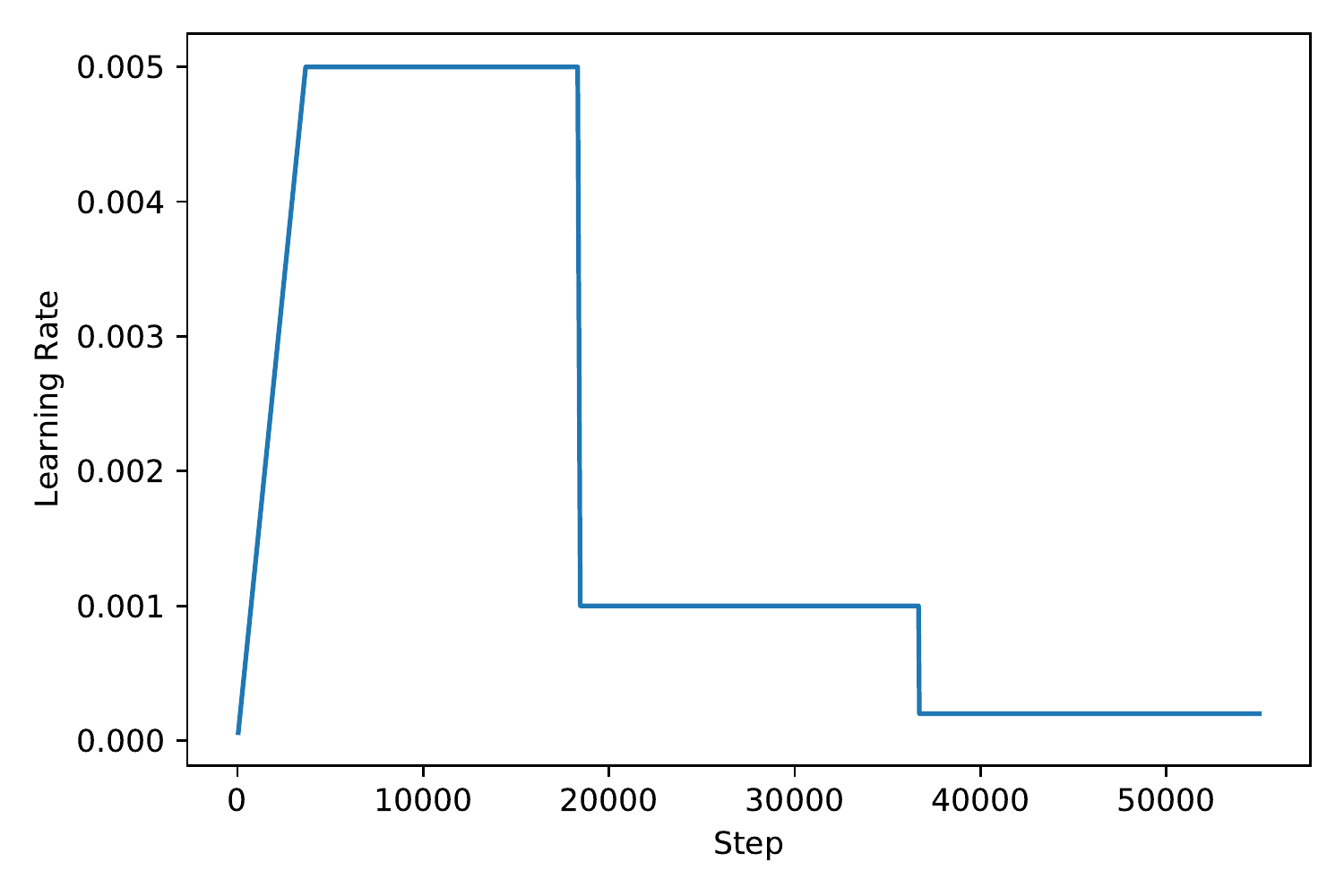}
    \caption{Learning rate for the model by iteration. The learning rate was adjusted using a stepped scheduler (\texttt{step size}$=2$, $\gamma=0.2$) and a warm-up (5000 iterations) to increase the rate at the beginning of training.}
    \label{fig:learning_rate}
\end{figure}

The model was optimized using \gls{sgd}, using an initial learning rate of 0.005, a momentum of 0.9, and weight decay of 0.0005.
The learning rate was adjusted over the course of the training using a stepped learning rate scheduler, which multiplied the rate by 0.2 every 5 epochs.
A warm-up period of 5,000 iterations was also implemented to ease into the initial learning rate.
The resulting learning rate curve is shown in \cref{fig:learning_rate}.

\section{Results}

\subsection{\Gls{exp_ntrain}: Number of training images}
\begin{figure}
    \centering
    \includegraphics[width=\linewidth]{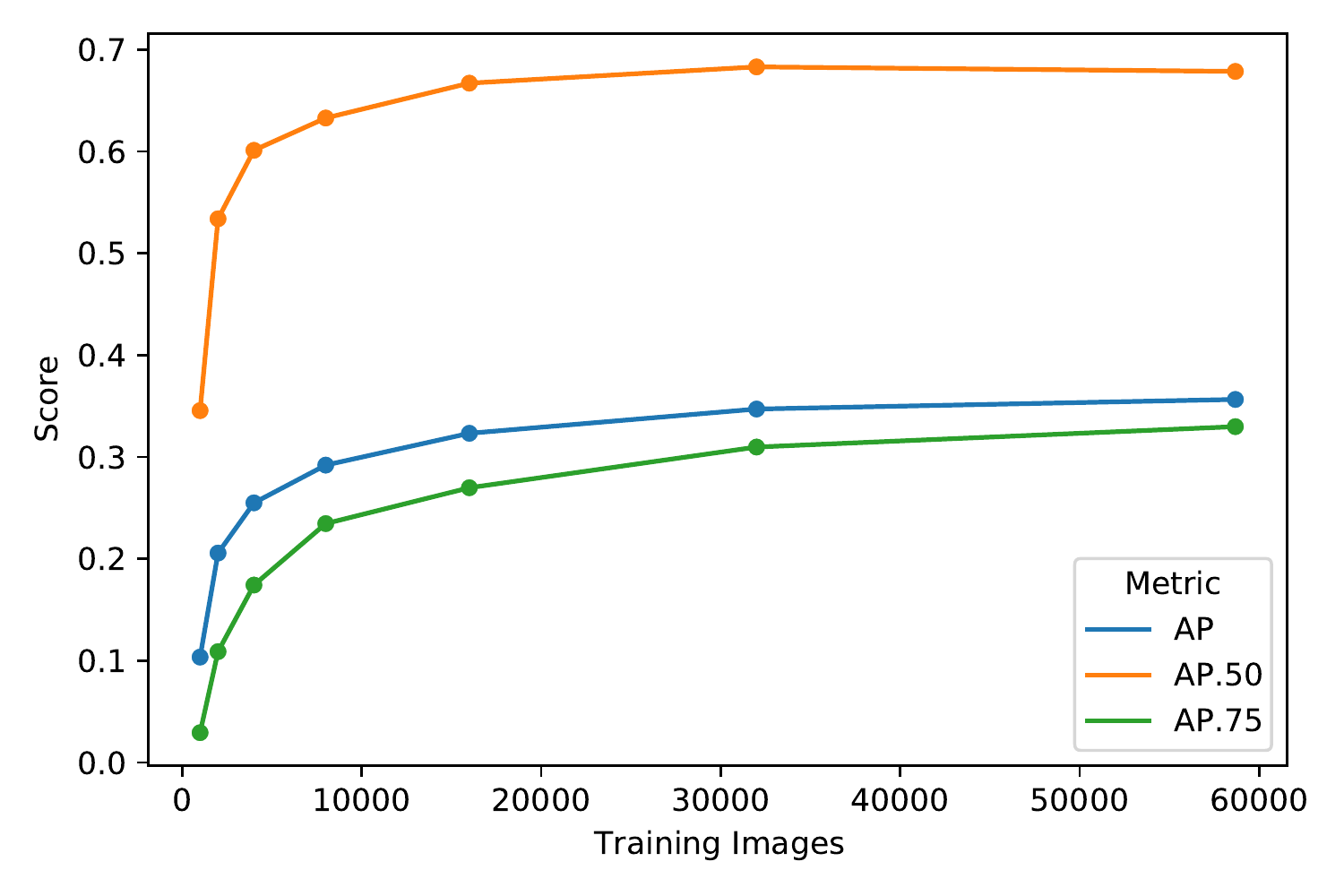}
    \caption{AP scores for networks trained using different numbers of training images in the \gls{sc} dataset and tested on the \gls{pa} dataset. The plot shows the 3 main AP measures (\gls{ap}, \gls{ap5}, and \gls{ap75}). The availability of more training images from \gls{sc} appears to improve the performance of the networks on the \gls{pa} testing set --- rapidly at first and then more gradually as the network sees tens of thousands of images in training.
    	At \n{} training images, \gls{ap}=0.36, \gls{ap5}=0.68, and \gls{ap75}=0.33.}
    \label{fig:ap_n_train}
\end{figure}

The first experiment tests how the performance of a \gls{nn} in detecting people in an unseen distribution of artistic images shifts depending on the number of stylized training images seen during fine-tuning.
\Cref{fig:ap_n_train} shows the performance of the network using the average precision metrics (\gls{ap}, \gls{ap5}, and \gls{ap75}) defined in the COCO \textit{detection evaluation metrics}\footnote{These metrics are used for all evaluation in this paper and are detailed on the COCO website at \url{https://cocodataset.org/\#detection-eval}}.

The precision rises sharply below about 10,000 images, but continue a slow improvement as more images are added to the training set.
On the \gls{ap5} metric, there is actually a small decrease in score between the \gls{nn} trained on 32,000 images and the one trained on \n{} images.
However, the \gls{ap} and \gls{ap75} scores continue to rise.

To understand why, it is important to consider the meaning of the three scores.
\gls{ap5} and \gls{ap75} are \acrlong{map} scores with different \gls{iou} thresholds and AP is the average of a series of scores at different \gls{iou} thresholds from .50 to .95.
The \gls{ap5} score reflects object detections that are less exact than the \gls{ap75} score.
So, it appears as though the network reaches a limit for moderately precise detections somewhere around 32,000 training images.
With more images, it is able to improve the precision of some detections while sacrificing some of the less precise bounding boxes.
Note that the decline in \gls{ap5} is relatively minor (0.4\%) compared to the rise in \gls{ap75} (2.0\%) between the networks trained on 32,000 and \n{} images.

\renewcommand\imggridwidth{0.19\textwidth}
\begin{figure*}[t]
     \centering
     \begin{subfigure}[b]{\imggridwidth}
         \centering
         \includegraphics[width=\textwidth]{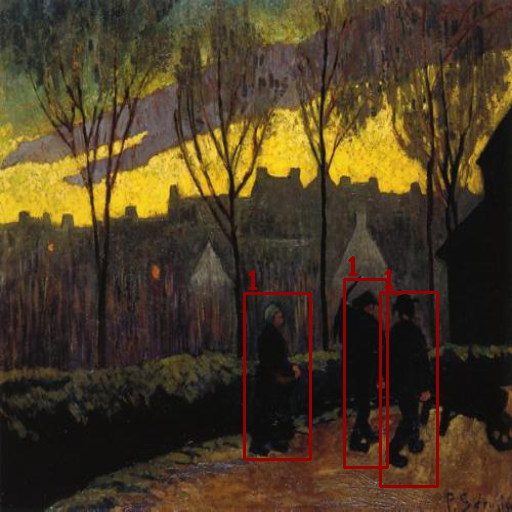}
         \caption*{Ground Truth}
     \end{subfigure}
     \hfill
     \begin{subfigure}[b]{\imggridwidth}
         \centering
         \includegraphics[width=\textwidth]{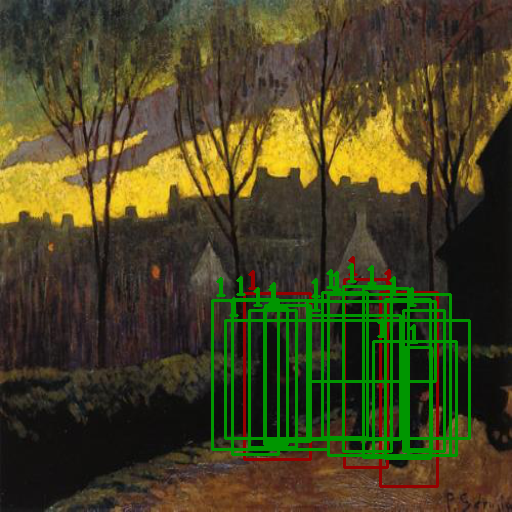}
         \caption*{1000}
     \end{subfigure}
     \hfill
     \begin{subfigure}[b]{\imggridwidth}
         \centering
         \includegraphics[width=\textwidth]{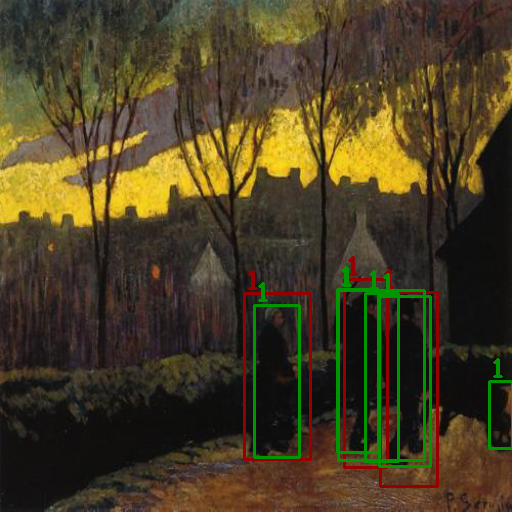}
         \caption*{8000}
     \end{subfigure}
     \hfill
     \begin{subfigure}[b]{\imggridwidth}
         \centering
         \includegraphics[width=\textwidth]{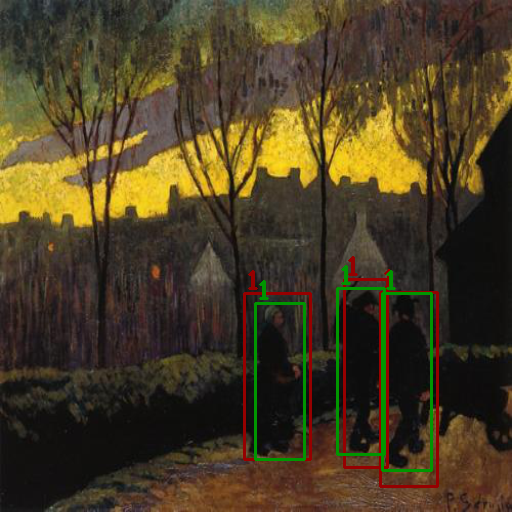}
         \caption*{32000}
     \end{subfigure}
     \hfill
     \begin{subfigure}[b]{\imggridwidth}
         \centering
         \includegraphics[width=\textwidth]{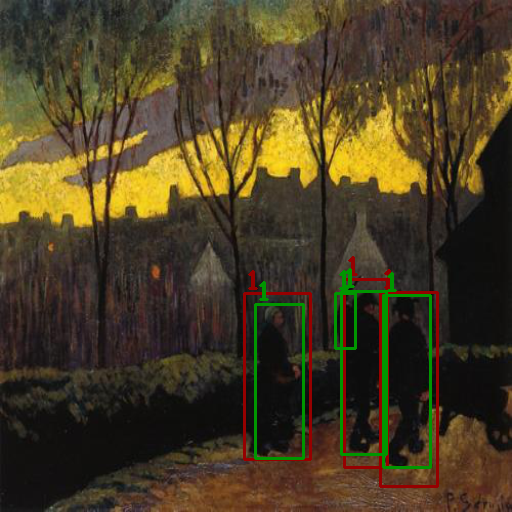}
         \caption*{\n{}}
     \end{subfigure}
     \hfill
        \caption{Ground truth (red bounding boxes) and results of person detection (green bounding boxes) with \gls{frrcnn} fine-tuned on \gls{sc} with increasing numbers of training images from 1000 to \n{}. Networks with fewer training images (left) produced far too many candidates. As the networks were able to train with a larger dataset, they narrowed their search and became more accurate overall. Note that in this example the network with 32,000 training images actually outperformed the network with \n{} training images as the latter network added an extra, unmatched candidate.}
        \label{fig:results_different_n}
\end{figure*}

\subsection{\Gls{exp_ap}: Performance on \gls{pa} dataset}

\Gls{exp_ap} tests the performance of the trained networks against the state of the art, as described in literature.
Four previous papers have published results for object detection on the \gls{pa} testing dataset using a variety of methods~\citep{Cai2015,Westlake2016,Redmon2016,Gonthier2020}.
Each reported the PASCAL VOC metric which corresponds to the AP.50 score used here, so a comparison is made based on this value.

\citet{Cai2015} and \citet{Westlake2016} are the creators of the \gls{pa} dataset.
\citet{Cai2015} used a \gls{dpm} as well as 2 R-CNN models pre-trained on ImageNet and fine-tuned on VOC 2012 and \gls{pa}.
\citet{Westlake2016} used Fast R-CNN models with a mix of backbones (CaffeNet, VGG1024, and VGG16) pre-trained on ImageNet and fine-tuned on either VOC 2007 or \gls{pa}, achieving the best result (AP.50=.58) with the VGG16 backbone fine-tuned on \gls{pa}.
A YOLO model was tested on the \gls{pa} dataset by \citet{Redmon2016} with mid-range success (AP.50=0.45) though this model was not fine-tuned on the \gls{pa} dataset nor was it trained or designed to recognize art specifically.
Finally, \citet{Gonthier2020} tested a number of \gls{mil} approaches in a \gls{wsod} task on the \gls{pa} dataset, achieving scores equal to those of \citet{Westlake2016} in their best models.

\begin{figure}[t]
    \centering
    \includegraphics[width=\linewidth]{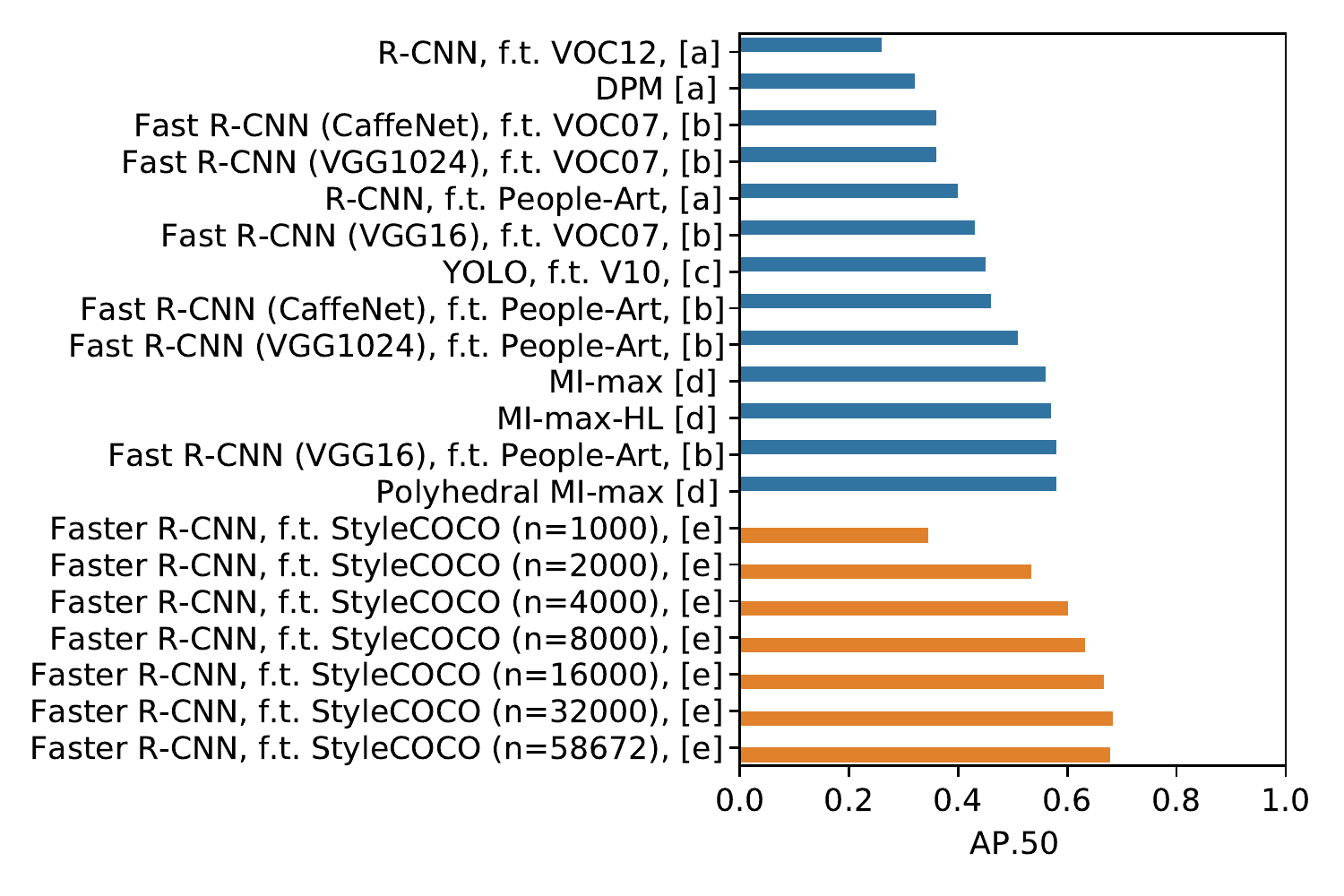}
    \caption{\Gls{ap5} scores on the \gls{pa} testing set for prior work (in blue) and results from this paper (in orange). Includes results from \citet{Cai2015} [a], \citet{Westlake2016} [b], \citet{Redmon2016} [c], \citet{Gonthier2020} [d] and this paper [e]. The method proposed here outperforms past detection efforts on the \gls{pa} dataset for every network trained with more than 4,000 images. The two networks trained with over 32,000 images achieve improvements of 10-percentage-points over the prior state of the art without having been trained on any of the images in the \gls{pa} dataset.}
    \label{fig:ap_compare}
\end{figure}

\begin{table}[t]
    \centering
    \caption{AP scores on the \gls{pa} dataset, best results}
    \begin{tabular}{cccc} \toprule
        \textbf{Paper}    & \textbf{\gls{ap}} & \textbf{\gls{ap5}} & \textbf{\gls{ap75}} \\ \midrule
        \citet{Cai2015} & --- & 0.4 & --- \\ 
        \citet{Westlake2016} & --- & 0.58 & --- \\ 
        \citet{Redmon2016} & --- & 0.45 & --- \\ 
        \citet{Gonthier2020} & --- & 0.58 & --- \\ 
        \textbf{This paper (n=\n)} & 0.36 & 0.68 & 0.33 \\ 
        \bottomrule
    \end{tabular}
    \label{tab:aps_from_others}
\end{table}

\begin{figure}[t]
    \centering
    \includegraphics[width=\linewidth]{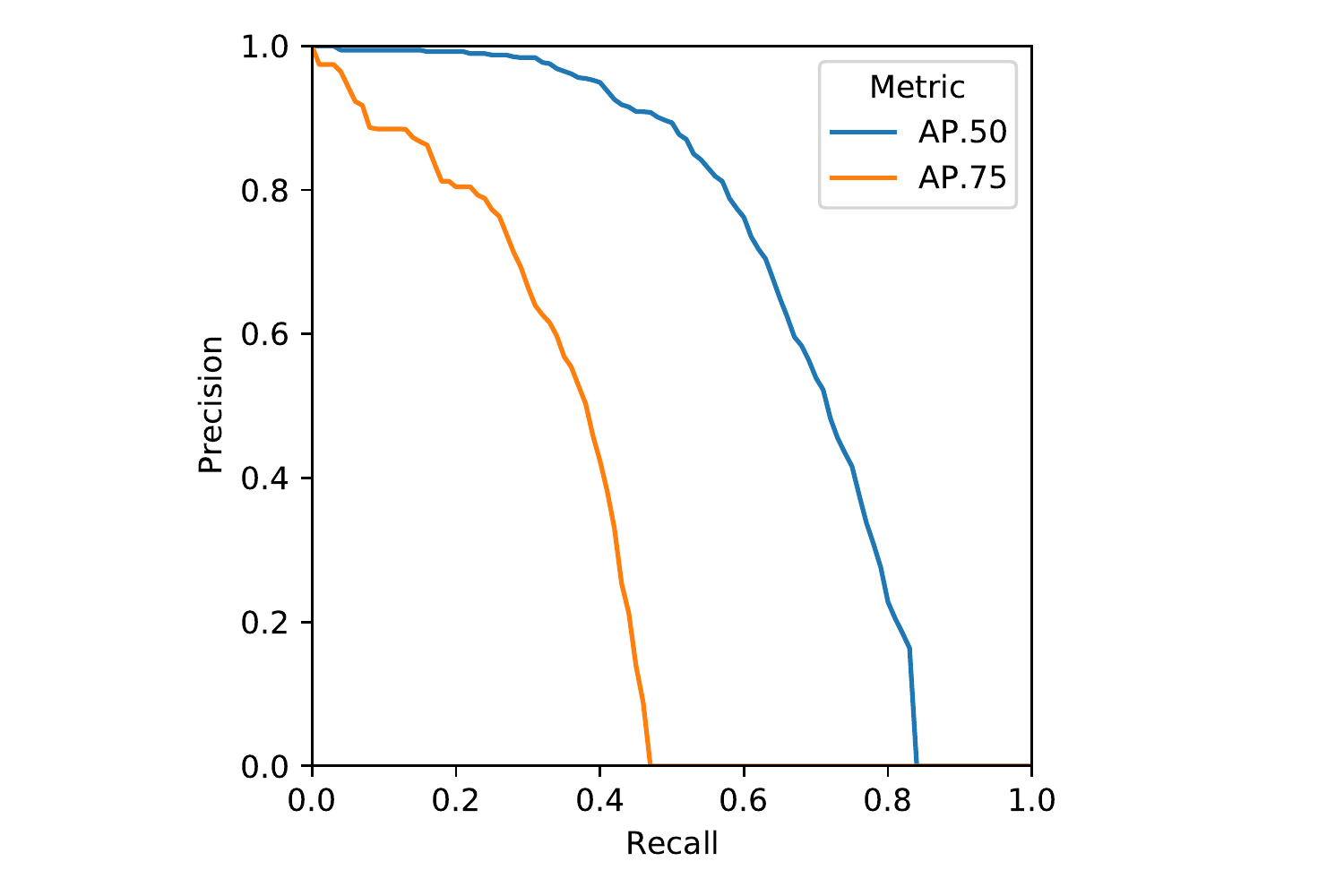}
    \caption{Precision-Recall curve for \gls{ap5} and \gls{ap75} metrics from evaluation on the \gls{pa} dataset for the \gls{frrcnn} network fine-tuned on \n{} images from \gls{sc}.}
    \label{fig:pr}
\end{figure}

The best result from each of the four papers is listed alongside our result in \cref{tab:pa_stats}; full results from those papers\footnote{Most of the papers attempted more than just one method and reported results from all of the trials.} and our tests in \gls{exp_ntrain} are plotted in \cref{fig:ap_compare}.
The approach proposed here achieves a full 10-percentage-point improvement over these methods, with an AP.50 score of $0.68$ on the \gls{pa} testing set, compared to the next best score of $0.58$.
For a more detailed picture of the performance, the precision-recall curve for the network trained with \n{} images is shown in \cref{fig:pr}.

\Cref{fig:results_images} shows some examples of the \gls{nn}'s range of performance on images from the \gls{pa} testing dataset.
The figure includes images where the \gls{nn} was inaccurate (A,B) as well as some where it failed to detect people (C,D).
In other images the \gls{nn} performed relatively well (E,F), labelling the correct number of people with reasonable accuracy, though it seemed to struggle to precisely label people who are occluded by other people.
In a final pair of images it correctly detected people that were unlabelled in the dataset (G,H), outperforming the humans that created the dataset.
In G the \gls{nn} correctly found a faint impression of a person in the left side of the image, though it appears to have been confused by the white space in the middle of the figure's body and has separately labelled the head, the lower body, and the entire person.
In H the \gls{nn} correctly found an additional head that was unlabelled in the right side of the image, but also mislabelled a mark at the top-right corner of the image.

\renewcommand\imggridwidth{0.24\textwidth}
\begin{figure*}
     \centering
     \begin{subfigure}[b]{\imggridwidth}
         \centering
         \includegraphics[width=\textwidth]{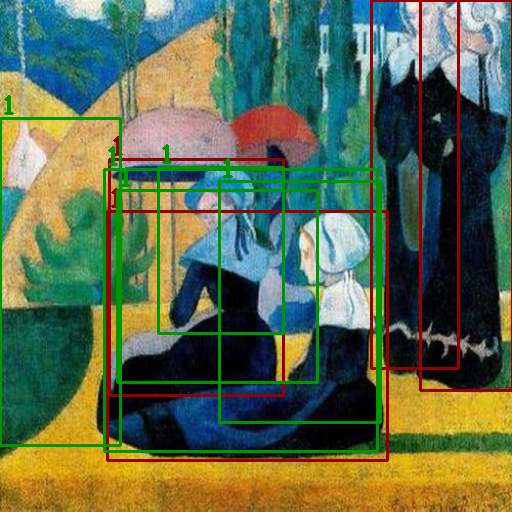}
         \caption*{A}
     \end{subfigure}
     \hfill
     \begin{subfigure}[b]{\imggridwidth}
         \centering
         \includegraphics[width=\textwidth]{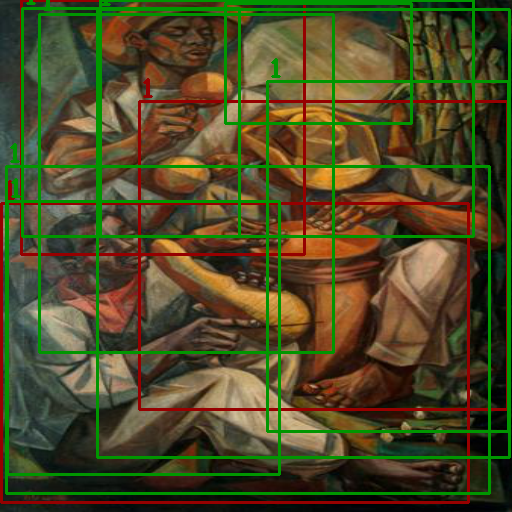}
         \caption*{B}
     \end{subfigure}
     \hfill
     \begin{subfigure}[b]{\imggridwidth}
         \centering
         \includegraphics[width=\textwidth]{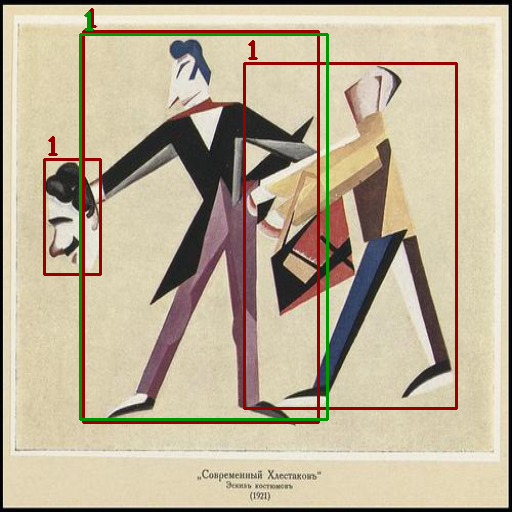}
         \caption*{C}
     \end{subfigure}
     \hfill
     \begin{subfigure}[b]{\imggridwidth}
         \centering
         \includegraphics[width=\textwidth]{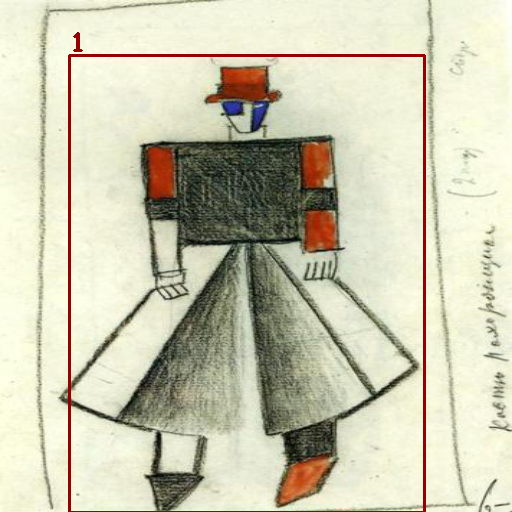}
         \caption*{D}
     \end{subfigure}
     \hfill
     \newline
     \begin{subfigure}[b]{\imggridwidth}
         \centering
         \includegraphics[width=\textwidth]{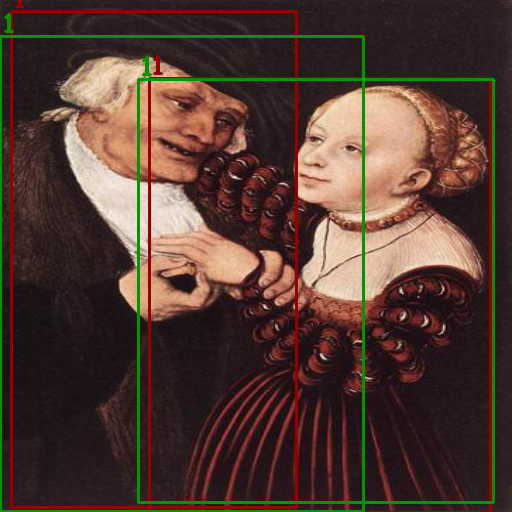}
         \caption*{E}
     \end{subfigure}
     \hfill
     \begin{subfigure}[b]{\imggridwidth}
         \centering
         \includegraphics[width=\textwidth]{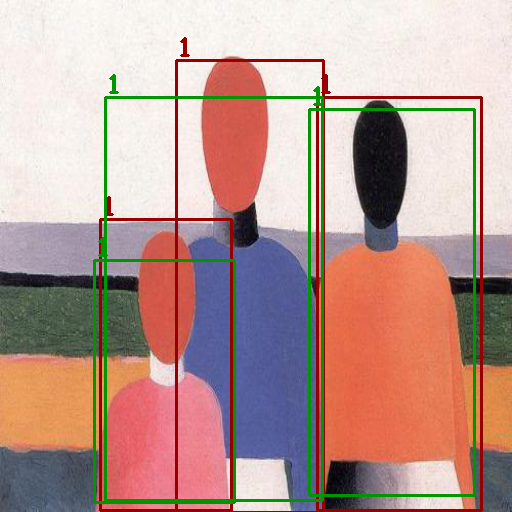}
         \caption*{F}
     \end{subfigure}
     \hfill
     \begin{subfigure}[b]{\imggridwidth}
         \centering
         \includegraphics[width=\textwidth]{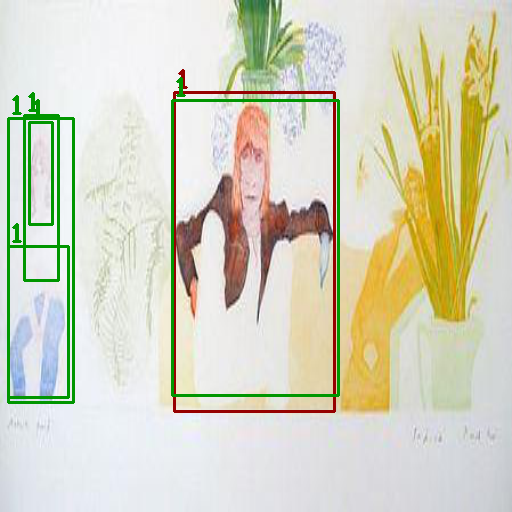}
         \caption*{G}
     \end{subfigure}
     \hfill
     \begin{subfigure}[b]{\imggridwidth}
         \centering
         \includegraphics[width=\textwidth]{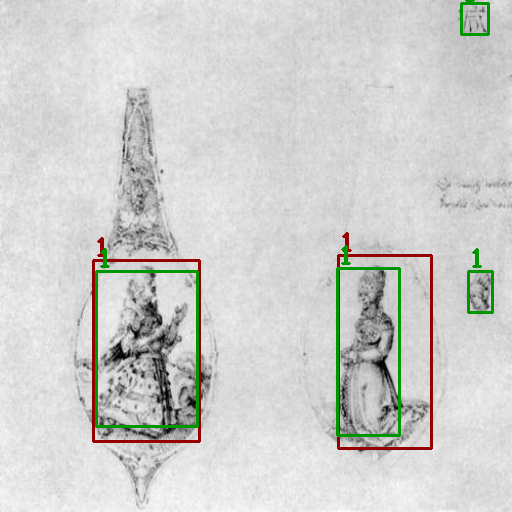}
         \caption*{H}
     \end{subfigure}
     \hfill
        \caption{Sample images from the \gls{pa} testing dataset with the ground truth (red bounding boxes) and people detected (green bounding boxes) using \gls{frrcnn} fine-tuned on \n{} images from \gls{sc}. In images A and B, the network did a poor job, failing to detect a number of people and placing wildly inaccurate bounding boxes for others in the busy scenes. Many of the people in images C and D were likely too abstract and the \gls{nn} failed to detect people in them, though it managed to quite accurately find the person in the middle of image C. Images E and F represent pretty good results. Each image has bounding boxes that are imperfectly placed, but both feature the correct number of detections with some bounding boxes that overlap the ground truth almost exactly. Finally, images G and H represent cases where the detector found people in the images that had not been labelled with a ground truth in the dataset, but that were correct nonetheless. These detections show how the \glspl{nn} can occasionally outperform humans at the task of detecting people in artwork. }
        \label{fig:results_images}
\end{figure*}

\section{Discussion}

\subsection{Failure Modes}
The leap in performance achieved with fine-tuning on \gls{sc} is itself useful, but it is just as interesting to examine the modes of failure of the method.
We have studied a variety of images to attempt to understand the \gls{nn}'s failings, and the examples in \cref{fig:results_images} illustrate some relevant issues.
Higher levels of abstraction such as that found in image D seem to confound the \gls{nn} --- especially in terms of the shape of the people.
It also failed to detect stick figures in another example image.

This is likely due to the way in which style transfer modifies images, largely shifting the colour and texture of the people in them without significantly warping their shapes.
This means that while the \gls{frrcnn} network is fine-tuned with people depicted in a wide range of colours and textures, it has not seen this combination of shapes as a person before and is therefore unable to recognize it as such.

Visual complexity appears to also pose a challenge for \gls{nn}.
In images A and B (\cref{fig:results_images}), the \gls{nn} returns some wildly inaccurate bounding boxes and outright misses some of the people in the image.
In B in particular, the resulting bounding boxes are dispersed almost randomly throughout the image.
B also points to another likely source of error in the data.

Given the well-documented~\citep{Buolamwini2018,Dulhanty2019} problem of gender, race, and age biases in machine learning, it is likely that this bias plays a role in the \gls{nn}'s failure on some images.
The \gls{frrcnn} network uses a backbone pre-trained on ImageNet which has documented imbalances in racial representation~\citep{Yang2020} that are shown to lead to biased models~\citep{Steed2020}.

The \gls{nn} presented here likely suffers a second type of bias related to artistic style.
The Painter-by-Numbers database used for style transfer in the creation of \gls{sc} draws most of its image set from \url{WikiArt.org}.
WikiArt catalogues around 250,000 artworks from more that 100 countries\footnote{\url{www.wikiart.org/en/about}}, but it appears to skew towards those living and educated in North America and Europe.
It is reasonable to assume, therefore, that an \gls{nn} trained on \gls{sc} would perform better, for example, on Renaissance paintings than on a print from the Japanese shin-hanga movement.
This was seen in testing as the \gls{nn} performed poorly on Kiyokata Kaburagi's \textit{At the Shore}.

Future work must consider these sources of bias --- the ImageNet backbone, the COCO base image dataset, and the artistic images used for style transfer --- and ways to identify these biases more specifically and mitigate its effects.
One way to more precisely identify the issues would be to test the \gls{nn} against the different image styles contained within the \gls{pa} dataset independently.
This could help to pinpoint the types of images that are difficult for the \gls{nn}.
Examination of saliency maps could also help to illuminate the source of difficulty in identifying people in specific artworks.

\subsection{Applications}
The technique in this paper can be applied to detect objects in non-photographic images, such as in museum collections and online databases.
This type of localization of people can be used to extract more detailed metadata such as the number and relative positions of people in art images.

There is an additional opportunity to extend this process beyond the recognition of people in artwork and into other object categories.
Using this technique it is possible to generate useful training data for all the object categories that are represented in the COCO dataset, thus circumventing one of the main limitations with existing datasets of non-photographic images, which include very few object categories (from 1 in People-Art to 7 in PACS).
\citet{Cai2015} note that objects do not appear with similar frequencies in artwork and people are over-represented, making it easier to collect databases of people in artwork, but harder to collect artwork depicting other minor categories.
If style transfer can be used to train object detectors for people in artwork using the COCO database as pre-labelled data, then it is likely possible to train detectors for the other 79 object categories in COCO as well.

Furthermore, the technique presented here could be used as a bootstrapping technique to build larger datasets of non-photographic images for object detection.
Using a \gls{nn} trained with the stylization technique one could annotate a large dataset (e.g. BAM), which holds 2.5 million images~\citep{Wilber2017}.
These bounding boxes could be verified and corrected using a crowdsourcing approach, resulting in a potentially very large dataset of non-photographic images which could be used to further improve object detection algorithms.

Other application areas for our approach are domains in which users can create their own content such as in video games, social networks, or internet forums.
In the game iNNk~\citep{Villareale2020}, an AI has to guess what object a person is drawing before the other players are able to do so.
For such systems it is challenging to obtain a large enough training set for accurate object detection, and they could therefore benefit from a dataset created using style-transfer.
Other games feature user-generated content that is moderated fully- or semi-automatically to remove or restrict drawn images that are disturbing or inappropriate for the audience.
A system trained on a style-transferred dataset could help to make these AI-driven processes more accurate and attuned to the styles of images produced in a particular game.

\subsection{Future Work}
In this initial exploration of the use of style transferred images for training object detectors in artwork, we used a general database of artistic images from the Painter-by-Numbers dataset to perform style transfer.
However, since the target set of artwork was known in advance it would have been possible to use that set instead.
A specialized version of \gls{sc} could be generated using unlabelled images from the target dataset as the style sources and that database could be used to train a \gls{nn} that is specially tuned to the styles of representation found in the target images.
This could yield additional improvements as the styled images in the training set more closely reflect the eventual targets.

Finally, though not used here, COCO also contains data for object segmentation, person keypoint detection and pose estimation, and detailed scene segmentation.
It would be interesting to see how well these could be translated to artwork using the style transfer method described here.

\bibliographystyle{IEEEtranN}
\bibliography{IEEEabrv,main.bib}

\begin{thebibliography}{26}
\providecommand{\natexlab}[1]{#1}
\providecommand{\url}[1]{#1}
\csname url@samestyle\endcsname
\providecommand{\newblock}{\relax}
\providecommand{\bibinfo}[2]{#2}
\providecommand{\BIBentrySTDinterwordspacing}{\spaceskip=0pt\relax}
\providecommand{\BIBentryALTinterwordstretchfactor}{4}
\providecommand{\BIBentryALTinterwordspacing}{\spaceskip=\fontdimen2\font plus
\BIBentryALTinterwordstretchfactor\fontdimen3\font minus
  \fontdimen4\font\relax}
\providecommand{\BIBforeignlanguage}[2]{{%
\expandafter\ifx\csname l@#1\endcsname\relax
\typeout{** WARNING: IEEEtranN.bst: No hyphenation pattern has been}%
\typeout{** loaded for the language `#1'. Using the pattern for}%
\typeout{** the default language instead.}%
\else
\language=\csname l@#1\endcsname
\fi
#2}}
\providecommand{\BIBdecl}{\relax}
\BIBdecl

\bibitem[LeCun et~al.(2015)LeCun, Bengio, and Hinton]{lecun_deep_2015}
Y.~LeCun, Y.~Bengio, and G.~Hinton, ``{Deep learning},'' \emph{Nature}, vol.
  521, no. 7553, pp. 436--444, may 2015.

\bibitem[Su et~al.(2019)Su, Vargas, and Sakurai]{su_one_2019}
J.~Su, D.~V. Vargas, and K.~Sakurai, ``{One Pixel Attack for Fooling Deep
  Neural Networks},'' \emph{IEEE Transactions on Evolutionary Computation},
  vol.~23, no.~5, pp. 828--841, oct 2019.

\bibitem[Cai et~al.(2015)Cai, Wu, Corradi, and Hall]{Cai2015}
H.~Cai, Q.~Wu, T.~Corradi, and P.~Hall, ``{The Cross-Depiction Problem:
  Computer Vision Algorithms for Recognising Objects in Artwork and in
  Photographs},'' may 2015, arXiv:1505.00110.

\bibitem[Boulton and Hall(2019)]{boulton_artistic_2019}
P.~Boulton and P.~Hall, ``{Artistic Domain Generalisation Methods are Limited
  by their Deep Representations},'' jul 2019, arXiv:1907.12622.

\bibitem[Ciecko(2020)]{ciecko2020}
B.~Ciecko, ``{AI Sees What? The Good, the Bad, and teh Ugly of Machine Vision
  for Museum Collections},'' in \emph{Museums and the Web 2020}.\hskip 1em plus
  0.5em minus 0.4em\relax Online: Museums and the Web, 2020.

\bibitem[Mollica(2018)]{mollica_send_2017}
J.~Mollica, ``{Send Me SFMOMA},'' in \emph{MW18: Museums and the Web 2018},
  Vancouver, BC, 2018.

\bibitem[Geirhos et~al.(2018)Geirhos, Rubisch, Michaelis, Bethge, Wichmann, and
  Brendel]{Geirhos2018}
R.~Geirhos, P.~Rubisch, C.~Michaelis, M.~Bethge, F.~A. Wichmann, and
  W.~Brendel, ``{ImageNet-trained CNNs are biased towards texture; increasing
  shape bias improves accuracy and robustness},'' nov 2018, arXiv:1811.12231.

\bibitem[Berry and Fagerjord(2017)]{berry_digital_2017}
D.~M. Berry and A.~Fagerjord, \emph{{Digital humanities}}.\hskip 1em plus 0.5em
  minus 0.4em\relax Cambridge, UK and Malden, MA: Polity Press, 2017.

\bibitem[Manovich(2018)]{manovich_ai_2018}
L.~Manovich, \emph{{AI Aesthetics}}.\hskip 1em plus 0.5em minus 0.4em\relax
  Moscow: Strelka Press, 2018.

\bibitem[Manovich(2020)]{manovich_cultural_2020}
------, \emph{{Cultural Analytics}}.\hskip 1em plus 0.5em minus 0.4em\relax
  Cambridge, MA: The MIT Press, 2020.

\bibitem[Merritt(2017)]{Merritt2017}
\BIBentryALTinterwordspacing
E.~Merritt. (2017, may) {Artificial Intelligence The Rise Of The Intelligent
  Machine}. American Alliance of Museums. [Online]. Available:
  \url{https://www.aam-us.org/2017/05/01/artificial-intelligence-
  the-rise-of-the-intelligent-machine/}
\BIBentrySTDinterwordspacing

\bibitem[Bailey(2019)]{Bailey2019}
\BIBentryALTinterwordspacing
J.~Bailey. (2019, apr) {Solving Art's Data Problem - Part One, Museums}.
  Artnome. [Online]. Available:
  \url{https://www.artnome.com/news/2019/4/29/solving-arts-data-
  problem-part-one-museums}
\BIBentrySTDinterwordspacing

\bibitem[Smith(2019)]{Smith2019}
\BIBentryALTinterwordspacing
J.~H. Smith. (2019, sep) {SMK's collection search levels up}. SMK Open.
  [Online]. Available:
  \url{https://medium.com/smk-open/smks-collection-search-levels-
  up-cf8e967e9346}
\BIBentrySTDinterwordspacing

\bibitem[Chang et~al.(2017)Chang, Amershi, and Kamar]{Chang2017}
J.~C. Chang, S.~Amershi, and E.~Kamar, ``{Revolt},'' in \emph{Proceedings of
  the 2017 CHI Conference on Human Factors in Computing Systems}, vol.
  2017-May.\hskip 1em plus 0.5em minus 0.4em\relax New York, NY, USA: ACM, may
  2017, pp. 2334--2346.

\bibitem[Deng et~al.(2009)Deng, Dong, Socher, Li, {Kai Li}, and {Li
  Fei-Fei}]{Deng2010}
J.~Deng, W.~Dong, R.~Socher, L.-J. Li, {Kai Li}, and {Li Fei-Fei}, ``{ImageNet:
  A large-scale hierarchical image database},'' in \emph{2009 IEEE Conference
  on Computer Vision and Pattern Recognition}.\hskip 1em plus 0.5em minus
  0.4em\relax IEEE, jun 2009, pp. 248--255.

\bibitem[Lin et~al.(2014)Lin, Maire, Belongie, Hays, Perona, Ramanan,
  Doll{\'{a}}r, and Zitnick]{Lin2014}
T.-Y. Lin, M.~Maire, S.~Belongie, J.~Hays, P.~Perona, D.~Ramanan,
  P.~Doll{\'{a}}r, and C.~L. Zitnick, ``{Microsoft COCO: Common Objects in
  Context},'' in \emph{Lecture Notes in Computer Science (including subseries
  Lecture Notes in Artificial Intelligence and Lecture Notes in
  Bioinformatics)}.\hskip 1em plus 0.5em minus 0.4em\relax Springer Verlag,
  2014, vol. 8693 LNCS, no. PART 5, pp. 740--755.

\bibitem[Wilber et~al.(2017)Wilber, Fang, Jin, Hertzmann, Collomosse, and
  Belongie]{Wilber2017}
M.~J. Wilber, C.~Fang, H.~Jin, A.~Hertzmann, J.~Collomosse, and S.~Belongie,
  ``{BAM! The Behance Artistic Media Dataset for Recognition Beyond
  Photography},'' in \emph{2017 IEEE International Conference on Computer
  Vision (ICCV)}, vol. 2017-Octob.\hskip 1em plus 0.5em minus 0.4em\relax IEEE,
  oct 2017, pp. 1211--1220.

\bibitem[Westlake et~al.(2016)Westlake, Cai, and Hall]{Westlake2016}
N.~Westlake, H.~Cai, and P.~Hall, ``{Detecting people in artwork with CNNs},''
  in \emph{Lecture Notes in Computer Science (including subseries Lecture Notes
  in Artificial Intelligence and Lecture Notes in Bioinformatics)}, vol. 9913
  LNCS.\hskip 1em plus 0.5em minus 0.4em\relax Springer Verlag, 2016, pp.
  825--841.

\bibitem[Redmon et~al.(2016)Redmon, Divvala, Girshick, and Farhadi]{Redmon2016}
J.~Redmon, S.~Divvala, R.~Girshick, and A.~Farhadi, ``{You only look once:
  Unified, real-time object detection},'' in \emph{Proceedings of the IEEE
  Computer Society Conference on Computer Vision and Pattern Recognition}, vol.
  2016-December.\hskip 1em plus 0.5em minus 0.4em\relax IEEE Computer Society,
  dec 2016, pp. 779--788.

\bibitem[Huang and Belongie(2017)]{Huang2017}
X.~Huang and S.~Belongie, ``{Arbitrary Style Transfer in Real-Time with
  Adaptive Instance Normalization},'' in \emph{Proceedings of the IEEE
  International Conference on Computer Vision}, vol. 2017-October.\hskip 1em
  plus 0.5em minus 0.4em\relax Institute of Electrical and Electronics
  Engineers Inc., dec 2017, pp. 1510--1519.

\bibitem[Gonthier et~al.(2020)Gonthier, Ladjal, and Gousseau]{Gonthier2020}
N.~Gonthier, S.~Ladjal, and Y.~Gousseau, ``{Multiple instance learning on deep
  features for weakly supervised object detection with extreme domain
  shifts},'' aug 2020, arXiv:2008.01178.

\bibitem[Buolamwini and Gebru(2018)]{Buolamwini2018}
J.~Buolamwini and T.~Gebru, ``{Gender Shades: Intersectional Accuracy
  Disparities in Commercial Gender Classification},'' in \emph{Proceedings of
  Machine Learning Research}, S.~A. Friedler and C.~W. Abstract, Eds.,
  vol.~81.\hskip 1em plus 0.5em minus 0.4em\relax New York, USA: PMLR, jan
  2018, pp. 1--15.

\bibitem[Dulhanty and Wong(2019)]{Dulhanty2019}
C.~Dulhanty and A.~Wong, ``{Auditing ImageNet: Towards a Model-driven Framework
  for Annotating Demographic Attributes of Large-Scale Image Datasets},'' may
  2019, arXiv:1905.01347.

\bibitem[Yang et~al.(2020)Yang, Qinami, Fei-Fei, Deng, and
  Russakovsky]{Yang2020}
K.~Yang, K.~Qinami, L.~Fei-Fei, J.~Deng, and O.~Russakovsky, ``{Towards fairer
  datasets},'' in \emph{Proceedings of the 2020 Conference on Fairness,
  Accountability, and Transparency}.\hskip 1em plus 0.5em minus 0.4em\relax New
  York, NY, USA: ACM, jan 2020, pp. 547--558.

\bibitem[Steed and Caliskan(2021)]{Steed2020}
R.~Steed and A.~Caliskan, ``{Image Representations Learned With Unsupervised
  Pre-Training Contain Human-like Biases},'' in \emph{Proceedings of the 2021
  ACM Conference on Fairness, Accountability, and Transparency}.\hskip 1em plus
  0.5em minus 0.4em\relax New York, NY, USA: ACM, mar 2021, pp. 701--713.

\bibitem[Villareale et~al.(2020)Villareale, Acosta-Ruiz, Arcaro, Fox, Freed,
  Gray, L{\"{o}}we, Nuchprayoon, Sladek, Weigelt, Li, Risi, and
  Zhu]{Villareale2020}
J.~Villareale, A.~V. Acosta-Ruiz, S.~A. Arcaro, T.~Fox, E.~Freed, R.~C. Gray,
  M.~L{\"{o}}we, P.~Nuchprayoon, A.~Sladek, R.~Weigelt, Y.~Li, S.~Risi, and
  J.~Zhu, ``{iNNk},'' in \emph{Extended Abstracts of the 2020 Annual Symposium
  on Computer-Human Interaction in Play}.\hskip 1em plus 0.5em minus
  0.4em\relax New York, NY, USA: ACM, nov 2020, pp. 33--37.

\end{thebibliography}

\end{document}